\documentclass{article}

 \usepackage[preprint]{lvl3m}

\usepackage[utf8]{inputenc} % allow utf-8 input
\usepackage[T1]{fontenc}    % use 8-bit T1 fonts
\usepackage{hyperref}       % hyperlinks
\usepackage{url}            % simple URL typesetting
\usepackage{booktabs}       % professional-quality tables
\usepackage{amsfonts}       % blackboard math symbols
\usepackage{nicefrac}       % compact symbols for 1/2, etc.
\usepackage{microtype}      % microtypography
\usepackage{xcolor}         % colors
\setcitestyle{comma,numbers,square}
\usepackage{graphicx}
\usepackage{multirow}
\usepackage{amsmath}
\usepackage{amssymb}
\usepackage{graphicx}
\usepackage{subcaption}

\title{LVLM-empowered Multi-modal Representation Learning for Visual Place Recognition}

\author{%
  Teng~Wang \\
  Department of Automation\\
  Southeast University\\
  Nanjing, Jiangsu 210019, China \\
  \texttt{wangteng@seu.edu.cn} \\
  \And
  Lingquan~Meng\\
  Department of Automation\\
  Southeast University\\
  Nanjing, Jiangsu 210019, China \\
  \texttt{menglingquan@seu.edu.cn} \\
  \And
  Lei~Cheng\\
  Department of Automation\\
  Southeast University\\
  Nanjing, Jiangsu 210019, China \\
  \texttt{leicheng@seu.edu.cn} \\
  \And
  Changyin~Sun\\
  Department of Automation\\
  Southeast University\\
  Nanjing, Jiangsu 210019, China \\
  \texttt{cysun@seu.edu.cn} \\
}

\begin{document}

\maketitle

\begin{abstract}
Visual place recognition (VPR) remains challenging due to significant viewpoint changes and appearance variations. Mainstream works tackle these challenges by developing various feature aggregation methods to transform deep features into robust and compact global representations. Unfortunately, satisfactory results cannot be achieved under challenging conditions. We start from a new perspective and attempt to build a discriminative global representations by fusing image data and text descriptions of the the visual scene. The motivation is twofold: (1) Current Large Vision-Language Models (LVLMs) demonstrate extraordinary emergent capability in visual instruction following, and thus provide an efficient and flexible manner in generating text descriptions of images; (2) The text descriptions, which provide high-level scene understanding, show strong robustness against environment variations. Although promising, leveraging LVLMs to build multi-modal VPR solutions remains challenging in efficient multi-modal fusion. Furthermore, LVLMs will inevitably produces some inaccurate descriptions, making it even harder. To tackle these challenges, we propose a novel multi-modal VPR solution. It first adapts pre-trained visual and language foundation models to VPR for extracting image and text features, which are then fed into the feature combiner to enhance each other. As the main component, the feature combiner first propose a token-wise attention block to adaptively recalibrate text tokens according to their relevance to the image data, and then develop an efficient cross-attention fusion module to propagate information across different modalities. The enhanced multi-modal features are simply compressed into the feature descriptor for performing retrieval.  Experimental results show that our method outperforms state-of-the-art methods by a large margin with significantly smaller image descriptor dimension. 
\end{abstract}

\section{Introduction}

Visual place recognition (VPR), also known as image geo-localization, refers to the task of determining whether a robot locates in a previously visited place by matching the local query image against a database of references with known geolocations. It is one of the essential problems in the field of robotics and computer vision, and finds great potentials in various applications, such as autonomous driving~\cite{Milford2012}, SLAM loop closure~\cite{Lowry2016}, and augmented reality~\cite{Shotton2013}. Despite great prospects, VPR is an extremely challenging task due to severe viewpoint variations, significant appearance changes caused by short-term (e.g., weather, lighting, etc.) and long-term (e.g. seasons, vegetation growth, etc.) environmental variation, and severe perceptual aliasing~\cite{Lowry2016} (multiple places in an environment may show high resemblance in appearance). 
\begin{figure*}[htbp]
         \centering
	\includegraphics[width = 1.0\textwidth, height=6.5cm]{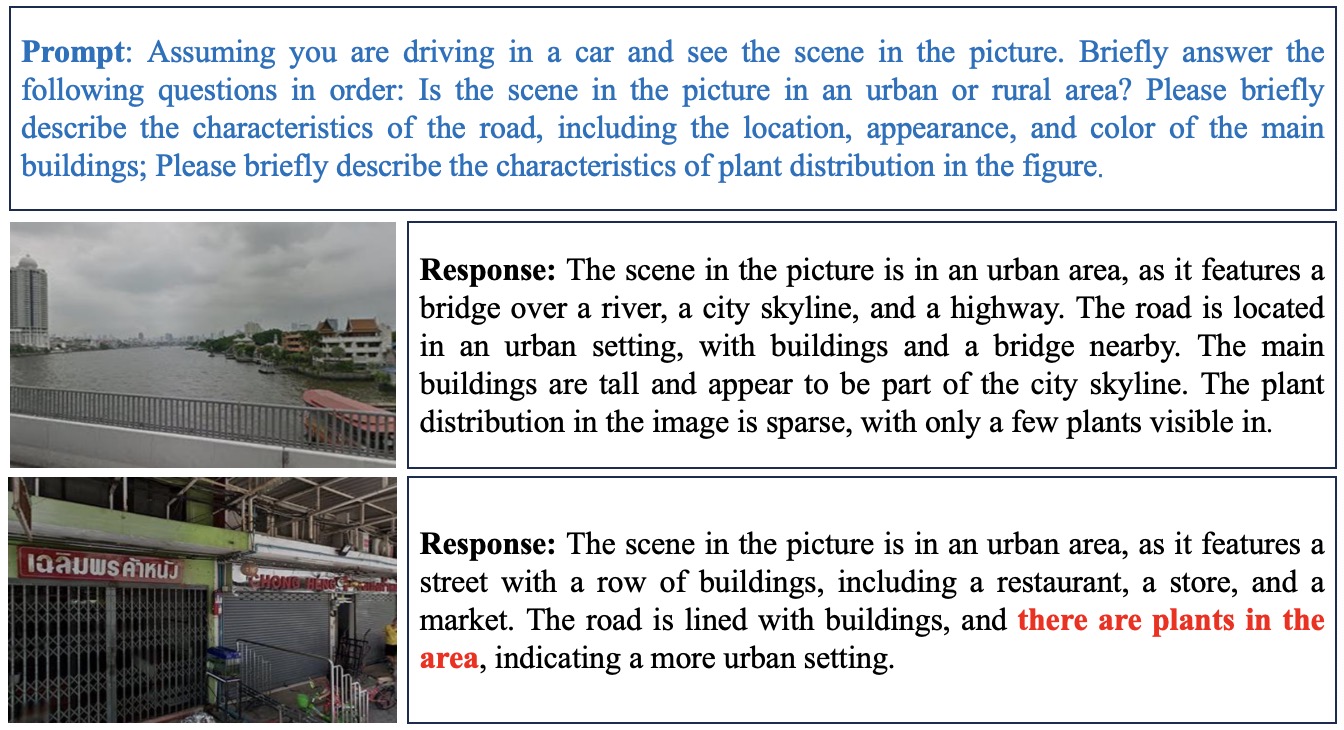}
	\caption{The visual understanding examples of LLaMA-Adapter V2~\cite{Gao2023}. When presented with a prompt, it could provide reasonable descriptions of the scene although some inaccurate details marked in red are also given. Besides, since language description entails identifying objects and their relative spatial relations in the scene, they show strong robustness against environmental variations. }
	\label{fig:Fig1-LLaMA-AdapterV2}
\end{figure*}

The VPR task is commonly approached as an image retrieval problem, where a global descriptor is generated to measure the similarity between the query and reference database images.  With the rapid development of deep learning, Convolution Neural Network (CNN) and Vision Transformer (ViT)~\cite{Dosovitskiy2021} have emerged as a powerful class of backbones for VPR and the global feature descriptors are obtained by aggregating these deep image features from backbone networks into a compact descriptor.  A variety of feature aggregation modules, such as NetVLAD~\cite{Arandjelovic2016, Wang2019, Nie2022, Kim2017}, GeM pooling~\cite{Berton2022,Lu2024,Radenovic2019}, and MLP-mixer~\cite{Ali-Bey2023} have been developed. These compact global representations facilitate fast place recognition with high accuracy. Unfortunately, satisfactory results cannot be achieved under severe appearance changes. Mainstream methods~\cite{Cao2020, Hausler2021, Wang2022,Lu2024ICLR} alleviate this issue by executing extra re-ranking, which performs geometric verification on the local features. These re-ranking techniques boost the accuracy of global retrieval at the expense of heavy computation and memory overhead, hindering the usage of these two-stage models for real-world large-scale VPR. 

On the other hand, recent advancements in Large Language Models (LLMs) have significantly stimulated the rapid development of Large Vision-Language Models (LVLMs), which leverage powerful Large Language Models (LLMs) as a brain to perform multimodal tasks. One of the extraordinary emergent capabilities of LVLMs is visual instruction following, as evidenced by MiniGPT-4~\cite{Zhu2024}, LLaVA~\cite{Liu2023} and LLaMA-Adapter V2~\cite{Gao2023}. These LVLMs opens up exciting avenues for the VPR research by providing textual descriptions of images without any extra human annotations. These language descriptions are desirable, since they could be instructed to provide high-level understandings of the visual scenes, which show strong robustness against environmental variations (e.g., weather, lighting, season, etc.) compared to raw pixel data (Fig.~\ref{fig:Fig1-LLaMA-AdapterV2}) . Motived by these observations, this work starts from a new perspective and attempts to leverage textual descriptions of a scene to enhance the global image features, aiming to build a discriminative holistic feature representation. Although promising, leveraging LVLMs to build multi-modal VPR solutions remains challenging in efficient multi-modal fusion. Furthermore, LVLMs will inevitably produce some inaccurate responses (Fig.~\ref{fig:Fig1-LLaMA-AdapterV2}), making the building of multi-modal VPR model even harder.

In view of the above consideration, we propose a novel \textbf{LVLM}-empowered \textbf{M}ulti-\textbf{M}odal \textbf{VPR} solution, dubbed \textbf{LVL3M-VPR} (Fig.~\ref{fig:framework}). This VPR model starts from leveraging LVLMs to generate language descriptions of images by delicately designing the prompts. The image and the associated description sentence are fed into their respective transformer encoders to generate domain-independent features. These multi-modal features are then fed into the feature combiner to take advantage of the complementarity between different modalities. As the main component, the feature combiner is composed of an attention-based text recalibration (AT-REC) module followed by a cross-attention multi-modal fusion (CA-MMF) network. The AT-REC is designed to guide the model to focus on those more informative and meaningful text tokens while filtering out those noisy ones. The  image and textual representations are then fused through CA-MMF to enrich each other.  Experimental results demonstrate that the proposed LVL3M-VPR achieve superior performance on benchmark datasets with significantly smaller image descriptor dimension. Main contributions of this work could be summarized as follows:
\begin{itemize}
 \item We leverage LVLM to build a novel multi-modal VPR solution based on transformer architecture. To the best of our knowledge, this is the first attempt to fuse visual features with textual features for VPR by introducing LVLM into the field of VPR. 
\item We build a novel multi-modal feature combiner, which follows the procedure of first filtering and then fusing. This combiner greatly enhances the representation power of the fused feature representations.
\item We conduct extensive experiments which confirm that our proposed LVL3M-VPR significantly outperforms the state-of-the-art on standard benchmark datasets with low memory requirement. 

\end{itemize}

\section{Related Works}

\textbf{Visual Place Recognition.} Nowadays, mainstream VPR research is dominated by deep learning techniques.  Global-retrieval-based (one-stage) VPR model is typically composed of a backbone network for feature extraction, followed by a trainable feature aggregation layer that compresses these deep features into a more powerful compact representation. The backbone networks evolve from CNN- based models~\cite{Arandjelovic2016, Wang2019, Nie2022, Kim2017, Hausler2021, Chen2014, Sunderhauf2015, Naseer2015, Chen2017,  Radenovic2019} to ViT-based models~\cite{Wang2022, Song2023, Li2024}. Particularly, the visual foundation models pre-trained on a large quantity of curated data, such as DINO v2~\cite{Oquab2024} are gaining increasing attention, and various adaptation techniques have been proposed to unleash the capability of these pre-trained models in the VPR task~\cite{Lu2024, Lu2024ICLR}.  As for feature aggregation, typical examples include NetVLAD~\cite{Arandjelovic2016} and its variants~\cite{Wang2019, Nie2022, Kim2017, Hausler2021}, R-MAC~\cite{Tolias2016}, the generalized mean (GeM)~\cite{Radenovic2019, Berton2022}, multi-scale attention aggregation~\cite{Wang2022}, and multilayer perceptron (MLP) based feature mixer~\cite{Ali-Bey2023}. These generated global deep descriptors demonstrate significant superiority over those handcrafted features. Nevertheless, they fail to achieve satisfactory performance under severe appearance variations. To enhance recognition accuracy, some works introduce extra re-ranking, which performs geometric verifications based on local features between the query and all initial candidates retrieved by global descriptor~\cite{Cao2020, Hausler2021, Wang2022,Lu2024ICLR}. These re-ranking techniques boost the accuracy of global retrieval at the expense of heavy computation and memory overhead. In addition, there are also several works that incorporate semantics~\cite{Naseer2017, Paolicelli2022},  depth~\cite{Piasco2021} and structure~\cite{Oertel2020} to enhance performance of VPR models under challenging environments.  Different from existing works, we for the first time introduce MLLMs into the field of VPR for generation high-level language descriptions of the visual scene, and attempt to build a discriminative global representation by fusing visual and text features. 

\textbf{Multi-modal Image Retrieval.} The research on multi-modal image retrieval is mainly concerned on the composed image retrieval (CIR) task~\cite{Vo2019}, where a query consists  of a reference image and a modification text that describes the desired changes. Most existing methods follows a fusion paradigm, where the multi-modal input pair is jointly embedded and compared against all candidate target images. Extensive research~\cite{Vo2019, Chen2020,  Baldrati2022, Lin2023, Liu2021, Liu2024} has been conducted on the fusion component, which is typically designed to combine the input modalities within the CLIP~\cite{Radford2021} or BLIP~\cite{Li2022} feature manifold. Among methods developed for multi-modal image retrieval, we are mostly related to VAL framework~\cite{Chen2020}, which also propose a transformer-based multi-modal fusion component. Please note that our work differs from the previous VAL in the following two main aspects: (1) our proposed multi-modal fusion components follows a paradigm of first filtering then combining, since the text descriptions from LVLM are noisy; (2) we leverage an efficient cross-modal attention to fuse multi-modal features while VAL performs self-attention on the concatenated visiolinguistic features. 

\section{Methodology}
\begin{figure*}[htbp]
        \centering
	\includegraphics[width = 0.98\textwidth]{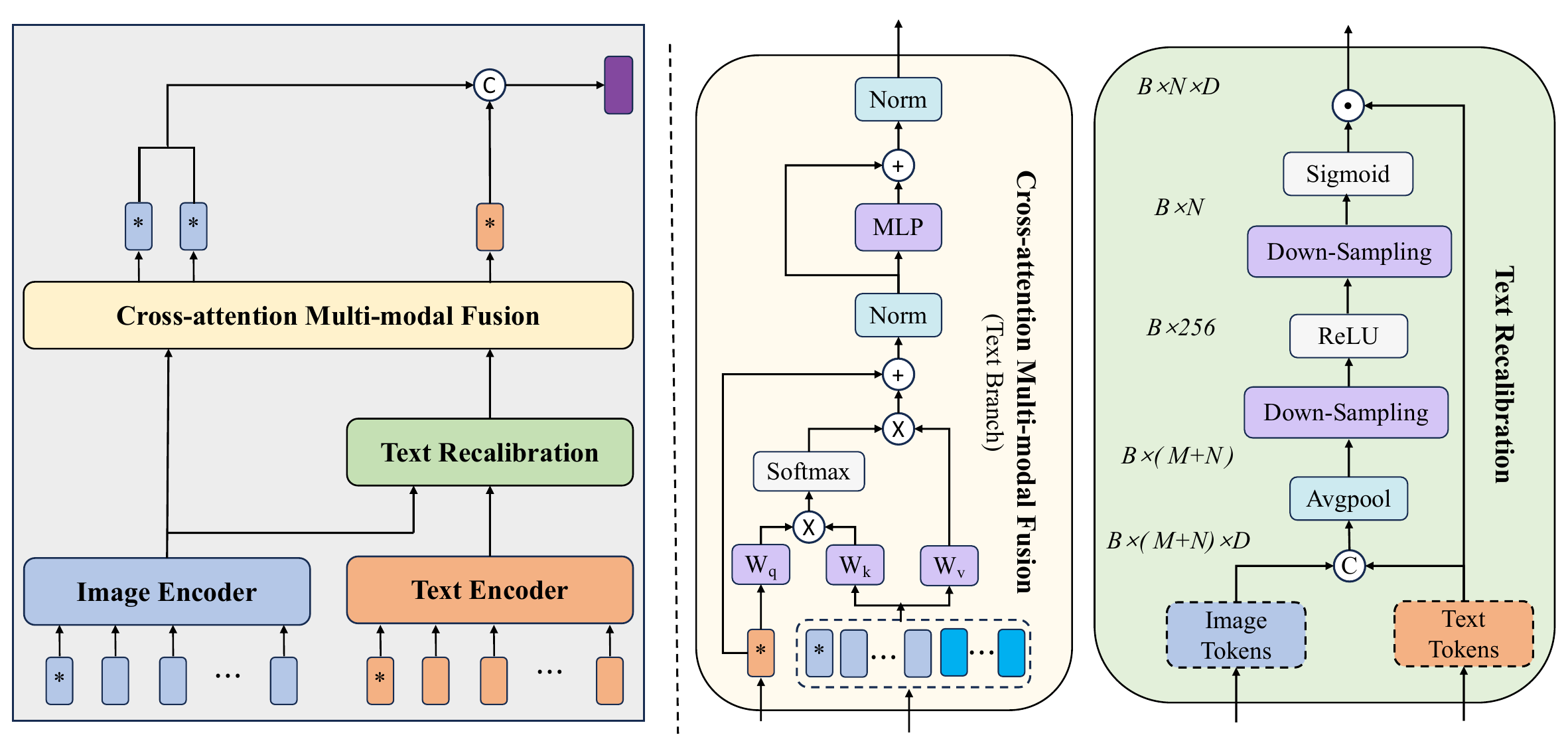}
	\caption{The overall framework of the proposed multi-modal VPR model. }
	\label{fig:framework}
\end{figure*}

Our goal is to leverage text description of the visual scene, which provide high-level semantic information of the scene and thus are more robust to environmental variations, to augment the performance of place recognition in challenging environments. To that end, we develop a novel LVLM-empowered multi-modal VPR model (LVL3M-VPR) to fuse visual and text information in an efficient manner (Fig.~\ref{fig:framework}). In this section, we describe the key components of the our LVL3M-VPR.

\textbf{LVLM Prompt.} LLaMA-Adapter V2~\cite{Gao2023} is chosen to generate text descriptions of the visual scene.  One of the most notable challenges we face when using  LLaMA-Adapter V2 is how to design a robust prompt to cope with the significant appearance variations for improving the performance of the pre-trained large models on the downstream VPR task. Since the visually salient static objects in a scene including man-made buildings and plants are critical to recognize a place, the prompt is designed to instruct the LLaMA-Adapter V2 to focus on these static objects as well as their relative spatial relations, as shown in Fig.~\ref{fig:Fig1-LLaMA-AdapterV2}. 

\textbf{Encoder.} Given the query image and the generated text description of the visual scene, we feed them into their respective transformer-based encoders to generate domain-independent features. For each branch, we append a learnable classification token ($\mathrm{CLS}$) at the beginning of the token sequence. To be specific, we employ the pre-trained visual foundation model DINOv2~\cite{Oquab2024} to extract image features by fine-tuning the last four layers of DINOv2 on VPR datasets for domain adaptation. The visual tokens from multi-levels of DINOv2 are summed up together to enrich the output image representations.  With respect to the text branch, we adapt the Bert encoder~\cite{Devlin2019} for VPR by adding a learnable linear layer on top of the frozen Bert encoder.   The output token sequences from the image and text branch are denoted as $\mathbf{X}=\left[\mathbf{x}_{\rm{cls}}, \mathbf{x}_1, \cdots, \mathbf{x}_{M-1}\right]$ and $\mathbf{Y}=\left[\mathbf{y}_{\rm{cls}}, \mathbf{y}_1, \cdots, \mathbf{y}_{N-1}\right]$, respectively.

\textbf{Attention-Based Text Recalibration.} It is well-known that LLMs suffer from the phenomenon of hallucination~\cite{Gunjal2024}. In addition, in a sentence, the noun words usually carry more information compared to those prep words. As a consequence, a text token selector is desired, aiming to filter out those noisy tokens while keeping the informative ones. In principle, the importances of text tokens could be measured by their relevances to the  image. We therefore develop a token-wise attention block, which adaptively recalibrates text tokens by explicitly modeling interdependencies between visual and text tokens. As depicted in Fig.~\ref{fig:framework}, the input image and text sequences are first concatenated along the patch dimension and then passed through an \textit{average pooling} operation to aggregate token features along the feature dimension, which allows the global token features to be used by all the subsequent layers. The pooling is followed by two fully-connected (FC) layers to capture dependencies between tokens. These FC layers produce  a set of per-token modulation weights by progressively reducing the dimensionality of the token descriptor to the length of the text sequence. The output of our attention block is obtained by rescaling the text sequence with these weights.The whole procedure is described as follows:
\begin{gather}
 \mathbf{Z} = \left[\mathbf{x}_{\rm{cls}}, \mathbf{x}_1, \cdots, \mathbf{x}_{M-1}, \mathbf{y}_{\rm{cls}}, \mathbf{y}_1, \cdots, \mathbf{y}_{N-1}\right],\\
 \mathbf{Z}' = {\rm Average \hspace{0.1cm}Pooling }(\mathbf{Z}) ,\\
 \mathbf{S} = \sigma(\mathbf{W}_{2} \sigma (\mathbf{W}_{1}\mathbf{Z}' )),\\
 \mathbf{Y}= \mathbf{Y}\odot \textbf{S},
\end{gather}
where $\sigma$ refers to the ReLU~\cite{Nair2010} function, $\mathbf{W}_1\in \mathbb{R}^{(M+N)\times T_{1}}$ and $\mathbf{W}_2\in \mathbb{R}^{T_{1}\times N}$; $\odot$ refers to the token-wise multiplication between the text token sequence and the weight vector.

\textbf{Cross-Attention Multi-Modal Fusion.} Recently, cross-attention is widely utilized for fusing multi-modal features~\cite{Vaswani2017, Rombach2022}, where each token from one sequence attends to all tokens from the other sequence. However, it requires $\mathcal{O}(M\times N)$ computations, where $M$ and $N$ refer to the lengths of the two input sequences, respectively. To reduce computations, we borrow idea from CrossViT~\cite{Chen2021}  and develop an efficient cross-attention fusion module, which leverages a query token as an agent to exchange information with other modality, as shown in Fig.~\ref{fig:framework}. For the image branch, to take advantage of both $\mathrm{CLS}$ token and patch tokens, we take the $\mathrm{CLS}$ token and an average of all patch tokens as two agents $\mathbf{z}^{m}$ and $\mathbf{z}^{a}$, which interacts with the sequence of tokens $\mathbf{Y}$ by attention separately to gather information from text data.  As for the text branch, we simply employ the $\mathrm{CLS}$ token as the query agent $\mathbf{z}^{t}$ since different text tokens are of different levels of importances. To convey more visual information to the text query token, we split the visual feature maps at three levels ($1\times1$, $2\times2$, and $3\times3$), and use \textit{average pooling}  to process patch features within the divided regions. The obtained 14 regional features $\mathbf{X}_a$ are appended to the original token sequence $\mathbf{X}$ for performing cross-attention computation.   Specifically, the fusion module in each branch is composed of $L$ cross-attention layers. At each layer $\ell \in \{1, 2, \cdots, L\}$, the query token $\mathbf{z}^{*}, *\in\{m, a, t\}$ enriches itself by interacting with token sequence from the other branch, which could be mathematically described as follows:
\begin{gather}
\mathbf{z}_{0}^{*} = \mathbf{z}^{*},  \mathbf{R}^{s} =\mathbf{R}^{a} = \mathbf{Y}, \mathbf{R}^{t} =  \left[\mathbf{X};\mathbf{X}_{a}\right],\\
 \mathbf{Q}_{\ell}^{*}=  \mathbf{z}_{\ell-1}^{*}\mathbf{W}^{Q, *}_{\ell} , \hspace{0.1cm} \mathbf{K}_{\ell}= \mathbf{R}^{*} \mathbf{W}^{K, *}_{\ell} , \hspace{0.1cm} \mathbf{V}_{\ell} = \mathbf{R}^{*} \mathbf{W}^{V, *}_{\ell}, \\
 \mathbf{\hat{z}}^{*}_{\ell} = {\rm MCA}(\mathbf{Q}^{*}_{\ell},  \mathbf{K}^{*}_{\ell}, \mathbf{V}^{*}_{\ell}) + \mathbf{z}_{\ell-1}^{*}, \\
 \mathbf{z}_{ \ell}^{*} = {\rm LN}({\rm MLP}({\rm LN}( \mathbf{\hat{z}}^{*}_{\ell})) +  \mathbf{\hat{z}}^{*}_{\ell}), \\
 \mathbf{z}^{*} =  \mathbf{z}_{L}^{*} ,
\end{gather}
where $\mathbf{W}^{Q, *}_{\ell}, \mathbf{W}^{K,*}_{\ell}, \mathbf{W}^{V,*}_{\ell} $ are learnable parameters; {\rm MCA} refers to the multi-head cross-attention operation in~\cite{Vaswani2017}; ${\rm MLP}$ and ${\rm LN}$ refer to multi-layer perceptron and layer normalization, respectively.  
%In this manner, the query agent takes advantage of both the \texttt{CLS} token and patch tokens, and thus encode more abundant information.Please note that in CrossViT~\cite{Chen2021}, the cross-attention fusion is followed by self-attention fusion, in which the learned information from the other branch by the class token is passed to its own patch tokens. In this manner, all the patches in one branch are updated by the patch information from the other branch. In comparison, only the class token is enhanced layer-by-layer in our method, which further reduces the computations.

\textbf{Final Image Representation.}  The query tokens $\mathbf{z}^{m}$ and $\mathbf{z}^{a}$ from the image branch and $\mathbf{z}^{t}$ from text branch are concatenated to construct the final image descriptor $\mathbf{f}$ for performing retrieval:
\begin{gather}
 \mathbf{f} =  \left[\mathbf{z}^{m}; \mathbf{z}^{a}; \mathbf{z}^{t}\right].
\end{gather}
%where $\left[;\right]$ refer to concatenation along the feature dimension. 

\section{Experiments}
\label{sec:experiments}

\subsection{Datasets and Evaluation Metrics.}
\textbf{Datasets.} We train our LVL3M-VPR on GSV-Cities~\cite{Ali-bey2022}, which is a large-scale dataset covering more than 40 cities across all continents over a 14-year period. The performance of the model is evaluated on three benchmark test datasets, including Pitts250k-test~\cite{Torii2013}, Mapillary Street-Level Sequences (MSLS)~\cite{Warburg2020}, and SPED~\cite{Zaffar2021}. For MSLS dataset, we evaluate LVL3M-VPR on both MSLS-Val and MSLS-challenge sets. The main characteristics of these three test benchmarks are summarized in Table~\ref{Tab:EvaluationBenchmarks}.

\begin{table*}[htbp]
	\caption{Summary of benchmark datasets for evaluation.. ``$+$'' indicates that the dataset includes the particular environmental variation,
and ``$-$'' is the opposite.}
	\label{Tab:EvaluationBenchmarks}
	\centering
	\resizebox{0.7\linewidth}{!}{
	\setlength{\tabcolsep}{1mm}{
	\begin{tabular}{l | ccccc | ll}
	\toprule
	\multirow{2}*{Dataset}   &  \multicolumn{5}{c|}{Variation}  & \multicolumn{2}{c}{Number}\\
		        \cline{2-8} 
			~                                                                     &Viewpoint  &Day/Night &Weather & Seasonal & Dynamic         & Database & Queries \\
			\midrule 
			Pitts250k-test~\cite{Torii2013}                       & $+$ & $-$ &  $-$ &$-$  & $+$                                                      &83,000         & 8,000 \\ 
			\midrule  
			MSLS-val~\cite{Warburg2020}                     & $+$ & $+$ &  $+$ &$+$  & $+$                                                   &18,871       & 740 \\
			MSLS-challenge~\cite{Warburg2020}          & $+$ & $+$ &  $+$ &$+$  & $+$                                                   &38,770       & 27,092 \\
			\midrule 
		        SPED~\cite{Zaffar2021}                                & $-$ & $-$ &  $+$ &$+$  & $-$                                                    &607             & 607 \\
       \bottomrule		 
     \end{tabular}}}
\end{table*}

\textbf{Evaluation Metrics.}  We follow standard evaluation procedure~\cite{ Arandjelovic2016, Ali-Bey2023,Warburg2020} to employ Recall@N (R@N) to evaluate the place recognition performance. Specifically, given a query image, it is regarded as “successfully localized" if at least one of the top-$N$ ranked images is within a threshold distance from the ground-truth location of the query. Default threshold values are used for all test datasets with 25 meters and $40^{\circ}$ for MSLS, 25 meters for Pitts250k-test and SPED. The percentage of query images which
have been correctly localized is reported as r@N. 

\subsection{Implementation Details.}
We fine-tune our LVL3M-VPR  on one NVIDIA GeForce RTX 4090 GPU using PyTorch. For the optimizer, we use the AdamW with the initial learning rate set as 6e-5 and reduced to 20\% of the initial value at the end of training procedure, for a maximum of 30 epochs. A training batch is composed of 120 places with 4 images for each place. The resolution of the input image is $224\times224$ and the embedding vectors from transformer-based backbones (ViT-B/14 for image branch, BERT-base-uncased for text branch)  are of length 768. To extract multi-scale visual features from input images, we sum up together the feature maps from the 2-nd layer, the 6-th layer, and the 12-th layer of DINOv2, corresponding to the low-level, middle-level, and high-level features respectively. In the cross-attention fusion component, the token-wise soft attention block is employed once and three layers of cross-attention fusion blocks are used for both text and image branches. For the  token-wise attention block, we set the length of the embedding vectors from the first FC layer to 256. At the same time, we set the token dimension from cross-attention fusion component to 768 as above. For the loss function, we follow previous works~\cite{Ali-Bey2023, Lu2024} to use Multi-Similarity loss to supervise the learning of the whole model~\cite{Wang2019MSL}.

\subsection{Comparison with State-of-the-Art Methods} 
In this section, we evaluate the performance of our proposed LVL3M-VPR  in visual place recognition on three challenging benchmarks by comparing it against several state-of-the-art (SOTA) VPR methods. Eight global-retrieval-based methods including GeM ~\cite{Arandjelovic2016}, NetVLAD~\cite{Arandjelovic2016}, SFRS~\cite{Ge2020}, CosPlace~\cite{Berton2022}, GCL~\cite{Vallina2023}, MixVPR~\cite{Ali-Bey2023}, EigenPlaces~\cite{Berton2023}, and CircaVPR~\cite{Lu2024} are considered. It is worth noting that we use the same training dataset GSV-Cities~\cite{Ali-bey2022} as the two latest works MixVPR~\cite{Ali-Bey2023} and CircaVPR~\cite{Lu2024}, whereas CosPlace~\cite{Berton2022} and EigenPlaces~\cite{Berton2023} are trained on their individually collected large-scale dataset SF-XL~\cite{Berton2022}.  In addition, we also consider two representative two-stage VPR methods TransVPR~\cite{Wang2022} and SelaVPR~\cite{Lu2024ICLR}, which requires extra re-ranking on local features to increase recall@N performance at the expense of heavy computation and memory overhead. Among these baselines, TransVPR,  CircaVPR and SelaVPR are transformer-based methods, which leverage ViT as backbones for feature extraction. Particularly, both CircaVPR and SelaVPR utilize the same pre-trained visual foundation model DINOv2~\cite{Oquab2024} for image feature extraction as our LVl3M-VPR. Different from our LVL3M-VPR, CircaVPR and SelaVPR develop various adaptation techniques to improve the performance of DINOv2 on the downstream VPR task. 

\begin{table*}[htbp]
	\caption{Comparison with existing SOTA methods on three benchmark test datasets. The best is highlighted in \textbf{bold} and the second is underlined. $\dagger$ denotes that the model is a two-stage method, which performs extra ranking on local features to improve the recall results at the expense of heavy computation and memory overhead.}
	\label{Tab:one-stage}
	\centering
	\resizebox{1.0\linewidth}{!}{
	\setlength{\tabcolsep}{0.5mm}{
	\begin{tabular}{l|c||ccc||ccc||ccc||ccc}
	\toprule
	\multirow{2}*{Model}  & \multirow{2}*{Dim} &  \multicolumn{3}{c||}{Pitts250k-test} & \multicolumn{3}{c||}{MSLS-val}  &  \multicolumn{3}{c||}{MSLS-challenge}  & \multicolumn{3}{c}{SPED}\\
		        \cline{3-14} 
			~& ~ & R@1 & R@5  & R@10 & R@1 & R@5  & R@10 & R@1 & R@5  & R@10 & R@1 & R@5  & R@10 \\
			\midrule   
			GeM ~\cite{Arandjelovic2016}                      & 2048        & 82.9 & 92.1 & 94.3            & 76.5 & 85.7 & 88.2      & - & -&-                                                    & 64.6 & 79.4& 83.5  \\
			NetVLAD~\cite{Arandjelovic2016}                & 32768      & 90.5 & 96.2 & 97.4            & 53.1 & 66.5 & 71.1      & 35.1 & 47.4 & 51.7                                 & 78.7 & 88.3& 91.4 \\
			SFRS~\cite{Ge2020}                                    & 4096        & - & - & -                             & 69.2 & 80.3 & 83.1       & 41.6 & 52.0 & 56.3                                 & - & - & -   \\
			CosPlace~\cite{Berton2022}                        & 512          & 89.7 & 96.4 & 97.7            & 82.8 &89.7  & 92.0       & 61.4 & 72.0 & 76.6                                  & 77.4 & 89.3& 92.6   \\
			GCL~\cite{Vallina2023}                                & 2048        & - & - & -                              & 79.5 & 88.1 & 90.1      & 57.9 & 70.7 & 75.7                                 & - & -& -  \\
			MixVPR~\cite{Ali-Bey2023}                         & 4096         & 94.6 & 98.3 & 99.0            & 88.0 & 92.7 & 94.6       & 64.0 & 75.9 &  80.6                                & 85.2 & 92.1 & 94.6   \\
			EigenPlaces~\cite{Berton2023}                   & 2048        & 94.1 & 97.9  & 98.7           & 89.1 & 93.8 & 95.0       & 67.4 & 77.1 & 81.7                                  & 69.9 & 82.9 & 87.6  \\
			CircaVPR~\cite{Lu2024}                              & 4096        & \textbf{95.1}& 98.5&99.2   & 90.0 & 95.4 & 96.4      & 69.0 & 82.1 & 85.7                                   & 88.3 & 94.2 & \textbf{95.4} \\
                         \midrule                   
                         %TransVPR (global)~\cite{Wang2022}           &  256         & - & - & -                              & 70.8 & 85.1 & 89.6      & 48.0  & 67.1 & 73.6                                   & - & - & -  \\
	 	        TransVPR$\dagger$~\cite{Wang2022}        &  -              & - & - & -                              & 86.8 & 91.2 & 92.4      & 63.9  & 74.0 & 77.5                                   & - & - &-    \\
                        % SelaVPR (global)~\cite{Lu2024ICLR}          & 1024        & - & - & -                               & 87.7 & 95.8 & 96.6      & 69.6  & \underline{86.9} & 90.1                                  & - & - &-    \\
                         SelaVPR$\dagger$~\cite{Lu2024ICLR}      &  -               & -& -& -                                & 90.8 & 96.4 & \textbf{97.2}      & 73.5  & \textbf{87.5} & \textbf{90.6 }          &-  & - &-   \\

      \midrule
                         LVL3M-VPR(ours)                                       &  2304        &\textbf{95.1}&\textbf{98.7} & \textbf{99.5}    &\textbf{92.0} & \textbf{96.4} & \underline{96.9}   & \textbf{73.8} & \underline{85.9} &\underline{88.1}   &\textbf{89.8} & \textbf{94.7} & \textbf{95.4}     \\
     \bottomrule		 
     \end{tabular}}}
\end{table*}

The quantitative results are shown in Table~\ref{Tab:one-stage}. It can be seen that, our LVL3M-VPR outperforms all other global-retrieval-based methods on all three benchmarks. For instance, on Pitts250-test, which shows significant viewpoint changes but no drastic environmental variations, LVL3M-VPR achieves competitive results with a slight increase at R@5 and R@10 compared to the second-performing CircaVPR. However, the dimensionality of our global image descriptor is only about half of that of CircaVPR. The benchmark MSLS is known to be challenging as it shows severe environment variations and includes some suburban or natural scene images, which lack salient landmarks for recognition and are thus prone to perceptual aliasing. On MSLS, performance is impressive with 92.0\% and 73.8\% achieved at R@1 on MSLS-val and MSLS-challenge, respectively. This is an absolute gain of 2.0\% and 2.9\% on MSLS-val respectively over CircaVPR and EigenPlace, which achieved 90.0\% and 89.1\% at R@1. Particularly, LVL3M-VPR outperforms the second-performing CircaVPR by a large margin of 4.8\% on MSLS-challenge. On SPED benchmark which features seasonal changes and day/night illumination, LVL3M-VPR improves CircaVPR and MixVPR by a margin of 1.5\% and 4.6\%, respectively. Furthermore, compared to the two-stage methods, our LVL3M-VPR outperforms both SelaVPR and TransVPR, with absolute
gains of 1.2\% and 5.2\%  on MSLS-val, as well as 0.3\% and 9.9\% on MSLS-challenge. All of the above observations demonstrate the effectiveness of our method on datasets presenting diverse and severe environment variations, including day/night illumination, weather and season changes. This could be attributed to the incorporation of text descriptions, which show strong robustness against environment variations.

\subsection{Qualitative Results} 
\begin{figure*}[htbp]
        \centering
	\includegraphics[width = 0.93\textwidth]{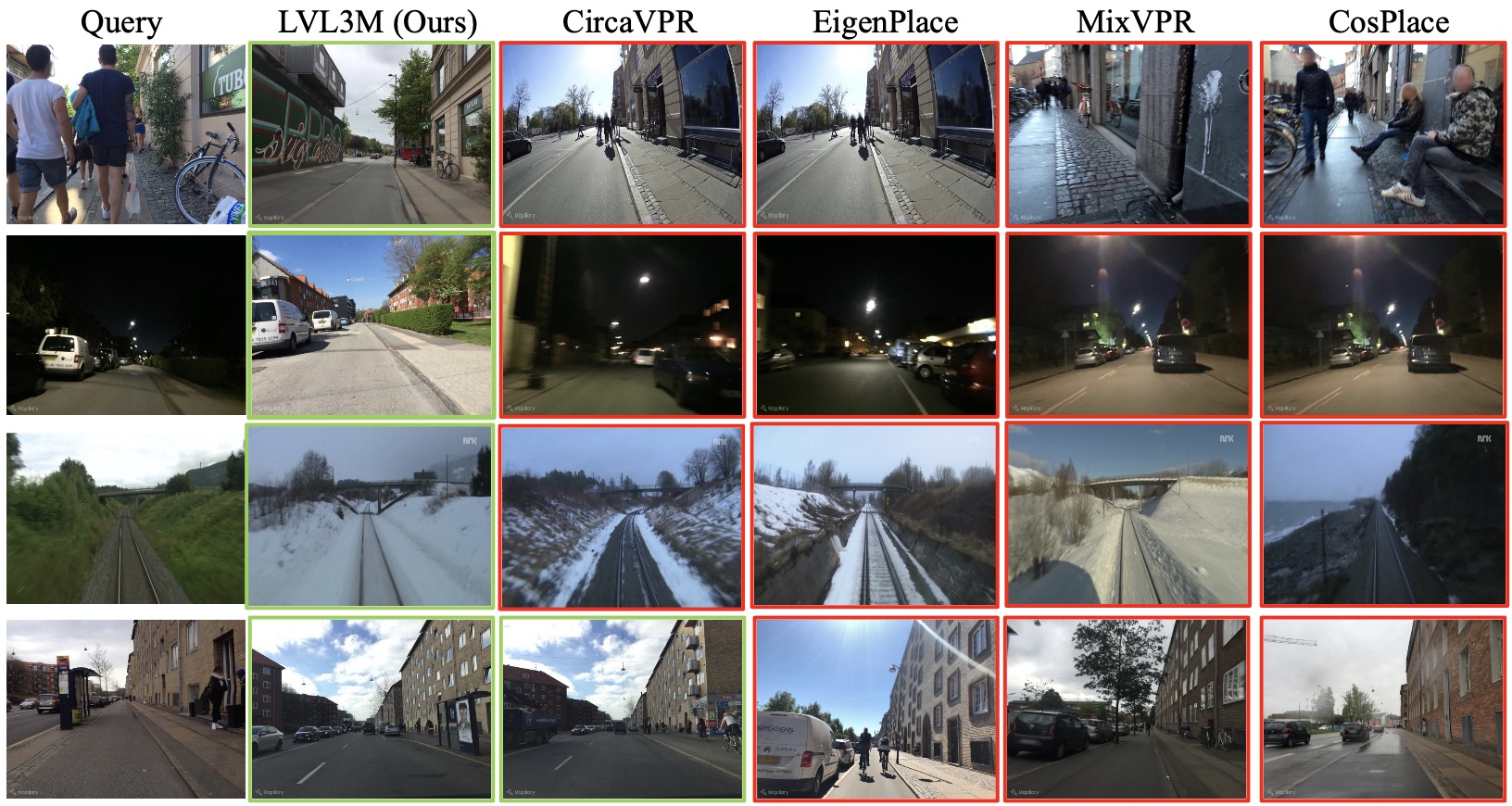}
	\caption{Comparison of retrieval results under challenging conditions. Red box and green box indicate a wrong and correct match, respectively.}
	\label{fig:Fig3-Retrieval-Results}
\end{figure*}

We give four examples to show the top-1 retrieved images of LVL3M-VPR under various challenging conditions in Fig.~\ref{fig:Fig3-Retrieval-Results}. In these four examples, significant appearance variations caused by dynamic objects, day/night illumination, seasonal change and viewpoint change occur between the query images and the reference images in database.  It can be seen that our LVL3M-VPR could successfully localize the matching database images for all the four cases. For instance, in the first example, walking pedestrians are included in the scene. Nevertheless, LVL3M-VPR succeed in retrieving the correct match by recognizing the structure and texture of the building as well the tree nearby. In the second example, the illumination change occurs where the query is taken during night and its reference is taken during the day. All the baseline methods retrieve night images from different locations. In contrast,  LVL3M-VPR retrieve the correct reference. In the third example, LVL3M-VPR retrieve the correct reference based most likely on the structure of the bridge and the curvature of the road while ignoring vegetation and snow on the ground. In the last example, both LVL3M-VPR and CircaVPR find the correct reference despite large viewpoint change. However,  LVL3M-VPR is capable of finding a closer reference based mostly on the bus stop board. These observations demonstrate the superiority of our model in these extremely challenging environments, which could be explained by its strong capability in capturing high-level context and fine details in the scene.

\subsection{Ablation Studies}
\textbf{Effectiveness of Main Components}. In LVL3M-VPR, three  main components are proposed, including attention-based text recalibration (AT-REC) module, cross-attention multi-modal fusion (CA-MMF) network, and the multi-scale feature fusion (MFF) strategy in DINOv2. We investigate the effectiveness of these three components by progressively removing them from our LVL3M-VPR and checking the changes in recall rates on the two benchmarks MSLS-val and SPED. The results are presented in Table~\ref{Tab:Ablation}. As can be seen, removing AT-REC from LVL3M-VPR leads to a decrease at R@1 from 92.0\% to 89.7\% (-2.3\%) on MSLS-val and from 89.8 to 89.3\% (-0.5\%) on SPED. This demonstrates the effectiveness of AT-REC in reducing ambiguity caused by the large vision-language model. Furthermore, we notice an absolute decrease of 2.0\% and 1.2\% at R@1 on MSLS-val and SPED respectively by continuing to remove the cross-attention fusion. Please note that when both AT-REC and CA-MMF are removed, we also employ the class tokens from both branches, combined with an average of patch tokens from image to construct the final retrieval descriptor for fair performance comparison. Finally, introducing multi-scale feature fusion within the DINOv2 brings an absolute gain of 1.5\% and 0.6\%  on MSLS-val and SPED respectively. This is reasonable, since both global context and fine details are critical to VPR under challenging conditions. 
\begin{table*}[htbp]
	\caption{Ablation study on the effectiveness of  the proposed three main components. ``+w/o'' denotes the specific component is removed from the model on the last row. } 
	\label{Tab:Ablation}
	\centering
	\resizebox{0.7\linewidth}{!}{
	\setlength{\tabcolsep}{2.5mm}{
	\begin{tabular}{c||ccc||ccc}
	\toprule
        \multirow{2}*{Model} & \multicolumn{3}{c||}{MSLS-val}  & \multicolumn{3}{c}{SPED}\\
        \cline{2-7} 
        ~ & R@1 & R@5  & R@10 & R@1 & R@5  & R@10  \\
	\midrule 
	LVL3M-VPR                            & 92.0 & 96.4  & 96.9    & 89.8 & 94.7 & 95.4  \\
        +w/o AT-REC                           & 89.7 & 95.3  & 95.8    &89.3  & 94.9 & 95.7 \\
        +w/o CA-MMF                         & 87.7 & 93.1  & 94.1    & 88.1 & 94.6 & 95.2 \\      
        +w/o MFF                                & 86.2 & 93.1  & 94.5    & 87.5 & 94.4  & 95.2 \\            
     \bottomrule		 
     \end{tabular}}}
\end{table*}

\textbf{In-depth Analysis of AT-REC}. In our AT-REC, the importances of text tokens are measured by their relevance to the image data, which is achieved by the token-wise attention block. We verify the superiority of the proposed attention block by constructing two variants for performance comparison.  The first variant is built by removing the image data from the inputs of the attention block, so that the importance of each text token is determined solely by its relevance to other text tokens, dubbed TextAtt. The second variant is to directly introduce a set of learnable parameters as the weight vector, dubbed LearnAtt. We re-train the two variants of LVL3M-VPR on GSV-Cities, and evaluate performances of the two variants on MSLS-val and SPED. As seen in Table~\ref{Tab:IndepthAnalysis}, our proposed AT-REC outperforms the two variants obviously on the two benchmark datasets, suggesting the necessity of leveraging image data to recalibrate text sequence. 

\begin{table*}[htbp]
	\caption{Ablation study on AT-REC and CA-MMF. For each component, different variations are considered and evaluated on MSLS-val and SPED. ``w/o'' denotes the technique is removed from the component, and``None'' denotes the component is completely removed from our LVL3M-VPR.}
	\label{Tab:IndepthAnalysis}
	\centering
	\resizebox{0.85\linewidth}{!}{
	\setlength{\tabcolsep}{2.5mm}{
	\begin{tabular}{c||c||ccc||ccc}
	\toprule
	\multirow{2}*{Component} & \multirow{2}*{Configuration} & \multicolumn{3}{c||}{MSLS-val}  & \multicolumn{3}{c}{SPED}\\
         \cline{3-8} 
        ~ &~ & R@1 & R@5  & R@10 & R@1 & R@5  & R@10  \\
	\midrule 
	 \multirow{4}*{AT-REC}     & None                           & 89.7 & 95.3  & 95.8    &89.3  & 94.9 & 95.7\\
          ~                                      &LearnAtt                       & 90.5 & 95.3 & 95.8          & 89.3 & 94.7 & 95.6 \\
	 ~                                      &TextAtt                           & 91.5 & 95.4 & 96.1          & 88.6 & 95.1 & 95.7\\
          ~                                      &ours                              & 92.0 & 96.4 & 96.9          & 89.8 & 94.7 & 95.4 \\
        \midrule                   
        \multirow{4}*{CA-MMF}   &None                             & 87.7  & 93.7 & 94.9            & 88.8 & 94.6 & 95.4 \\
         ~                                     & w/o AvePatToken         & 91.1  & 96.0  & 96.6           & 88.9 & 94.7 & 95.7 \\
         ~                                     &w/o MsPatToken           & 91.0  & 95.0  & 96.1           & 89.1& 94.6 & 95.6 \\
         ~                                     & ours                              &92.0 & 96.4.   & 96.9          & 89.8 & 94.7 & 95.4 \\
      \bottomrule		 
      \end{tabular}}}
\end{table*}

\begin{figure*}[htbp]
        \centering
	\includegraphics[width = 0.92\textwidth]{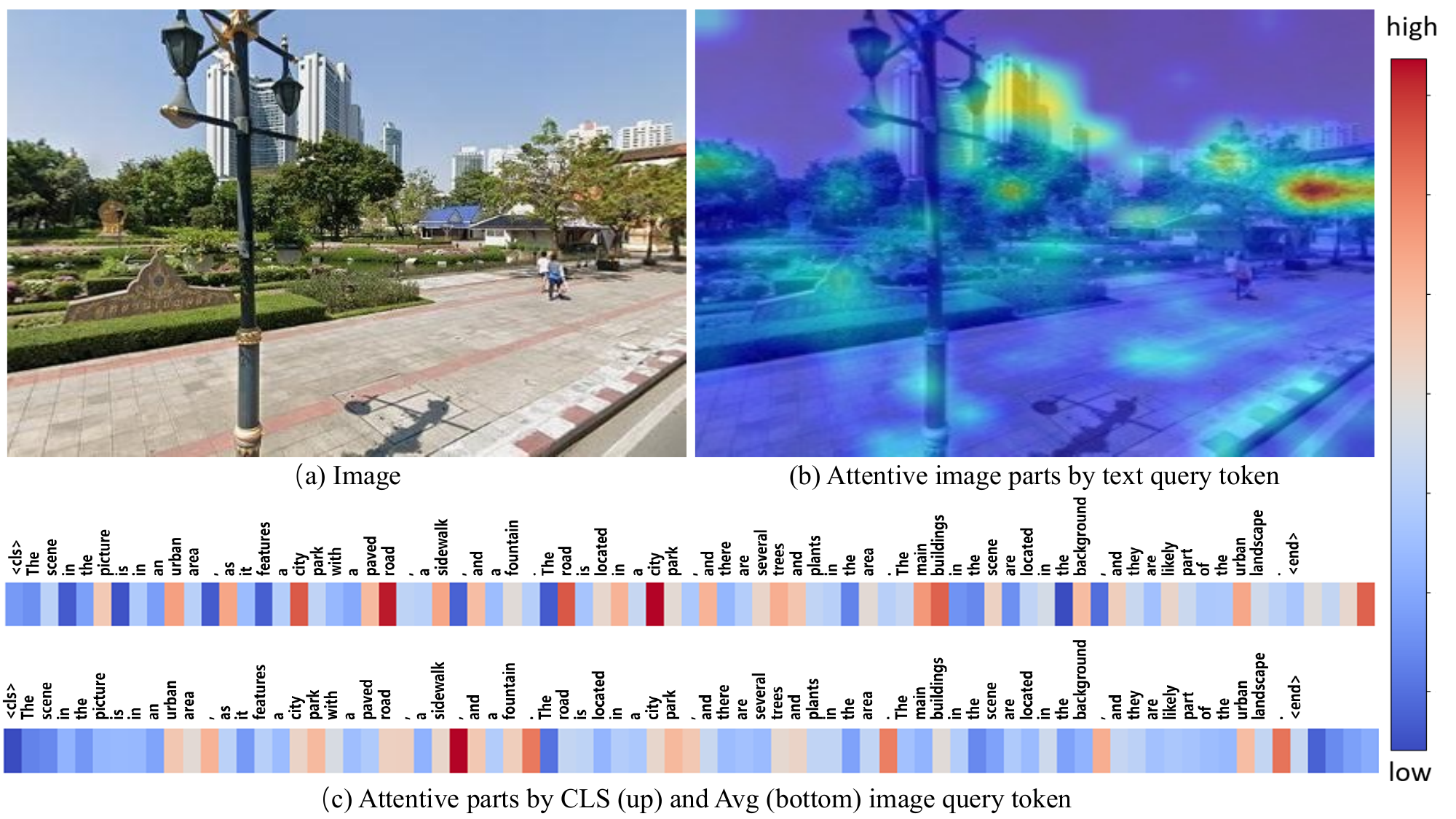}
	\caption{Visualization of the learned bi-directional attention maps within CA-MMF component. }
	\label{fig:Fig5-Cross-View-Visualization}
\end{figure*}

\textbf{In-depth Analysis of CA-MMF}. In the proposed CA-MMF, multiple techniques have been proposed to enhance fine details in the feature descriptor. First, for the image branch, an average of patch tokens from image is incorporated to construct the image query token, which interacts with the text token sequence by attention to gather useful information from text data. Second, for the text branch, the average pooled features from multi-scale regions of images are considered for interacting with the text query token in order to convey more visual features to the text branch. We verify the effectiveness of the two techniques, dubbed AvePatToken and MsPatToken, by respectively removing them from the cross-attention fusion module and checking the performance changes. It can be seen from Table~\ref{Tab:IndepthAnalysis} that removing AvePatToken reduces R@1 by a margin of 0.9\%  and 0.9\% on MSLS-val and SPED, respectively. It is worth mentioning that when the AvePatToken is removed, the dimensionality of the retrieval feature descriptor is reduced to 768*2.  Similar observations could be made from removing the MsPatToken. Furthermore, we gain an obvious decrease of 4.3\% and 1.0\% at R@1 on MSLS-val and SPED respectively by removing the entire CA-MMF component from LVL3M-VPR. These observations demonstrate the superiority of the proposed CA-MMF in enhancing multi-modal feature representation learning. Furthermore, to gain a deeper insight into how CA-MMF works, we visualize the learned attention maps involved in the bi-directional cross-attention module in Fig.~\ref{fig:Fig5-Cross-View-Visualization}. It can be seen that the query token from the text branch mainly attends to those salient objects in the visual scene, thus enriching itself with rich fine details included in the visual data. On the other hand, the $\mathrm{CLS}$ query token from the image branch mainly focus on those meaningful nouns, such as ``city'', ``road'', ``buildings'', and ``trees'', whereas the average token focus on the words including ``sidewalk'', ``foundation'' and ``park''. 

\section{Conclusion and Limitation}
\label{sec:conclusion}
In this work, we take advantage of the strong power of LVLM in visual instruction following, and for the first time build a multi-modal VPR solution which complements the visual data by the text description of the visual scene. To fuse multi-modal features in an efficient manner, we propose a novel multi-modal combiner, which follows the procedure of first filtering and then fusion.  To be specific, the text sequence is first recalibrated by a token-wise attention block, which measure the relevance between each tex token and image data to determine its importance. Then, the features from different modalities are fused through an efficient cross-attention fusion module, where special query tokens are constructed for each modality which serve as agents for interacting with tokens from the other modality for information exchange.  Extensive experiment results on benchmark datasets demonstrate the our proposed methods outperform previous SOTA obviously while maintaining much smaller feature descriptor dimension, which further suggest the usefulness of text descriptions in enhancing VPR under challenging conditions.

Although some improvements are achieved,  our method is still limited by the performance of the employed vision-language model LLaMA-Adapter V2. When presented with the specifically designed prompt, LLaMA-Adapter V2 fails to produce accurate responses with high confidence. We speculate this might be due to the following two facts: (1) The prompt needs to the improved. The VPR datasets features high diversity in viewpoint, environment, and appearance. For instance, places from urban and natural environments show quite different distribution in plants and man-made infrastructures. The prompt we have designed could only cope with these variations to some extent. (2) LLaMA-Adapter V2 itself has some difficulty in understanding complex scenes. In future, we will improve the design of prompt while attempt some other more powerful VLVM.

%%%%%%%%%%%%%%%%%%%%%%%%%%%%%%%%%%%%%%%%%%%%%%%%%%%%%%%%%%%%
\newpage
\appendix

\section{Final Feature Descriptor for Retrieval}
In LVL3M-VPR, we employ the query tokens from both the image and text branches to construct the final feature descriptor for retrieval. To verify the necessity of each component, we consider the following two alternatives. The first one is to employ the query tokens (including the $\mathrm{CLS}$ and average tokens) from the image branch to build the retrieval feature descriptor, denoted as IM-$\mathrm{CLS}$+Avg. The last one is to simply employ the $\mathrm{CLS}$ query tokens from the text branch as the final feature descriptor, denoted as TX- $\mathrm{CLS}$. The recall rates on benchmarks MSLS-val and SPED under the two different settings are presented in Table~\ref{Tab:featuredescriptor}. It is observed that only employing the query tokens from the image branch achieves R@1 of 90.0\% and 87.6\% respectively on MSLS-val and SPED datasets. Incorporating query token from the text branch brings an absolute increase of 2.0\% and 2.2\% at R@1 respectively on MSLS-val and SPED although the feature dimensionality is increased from $768*2$ to $768*3$, which is still much smaller than that of previous SOTA  method Circa~\cite{Lu2024} (i.e., 2304 vs 4096). This observation also suggests the query token features from different modalities complement each other. Therefore, we prefer to concatenate the three query tokens from both branches to build the final feature descriptor.
\begin{table*}[htbp]
	\caption{Ablation study on retrieval feature descriptor.}
	\label{Tab:featuredescriptor}
	\centering
	\resizebox{0.75\linewidth}{!}{
	\setlength{\tabcolsep}{1.5mm}{
	\begin{tabular}{c||c||ccc||ccc}
	\toprule
	\multirow{2}*{Setting} & \multirow{2}*{Dim} & \multicolumn{3}{c||}{MSLS-val}  & \multicolumn{3}{c}{SPED}\\
         \cline{3-8} 
        ~ &~ & R@1 & R@5  & R@10 & R@1 & R@5  & R@10  \\
	\midrule 
     	 IM-$\mathrm{CLS}$+Avg            & $768*2$                        & 90.0 & 95.1 & 96.2          & 87.6 & 94.6 & 95.6 \\
	 TX- $\mathrm{CLS}$                   & $768$                           & 84.6 & 91.2 & 93.2          & 84.2 & 93.3 & 94.4 \\
	 ours                                             &$768*3$                         &92.0 & 96.4.   & 96.9        & 89.8 & 94.7 & 95.4 \\
       \bottomrule		 
      \end{tabular}}}
\end{table*}

\section{Number of Cross-attention Fusion Layers}
To demonstrate the effectiveness of fusing multi-modal features by using cross-attention modules, we use different number of cross-attention layers, i.e., 1,3, 5, 7, and report the recall rates on benchmarks MSLS-val and SPED in Table~\ref{Tab:crossattentionlayer}. An absolute gain of 1.5\% at R@1 is obtained on MSLS-val when $L$ is increased from 1 to 3. However, when $L$ is further increased from 3 to 5, we seen an obvious decrease of 1.5\% at R@1 on MSLS-val. In comparison, on SPED, changing the number of cross-attention layers does not bring obvious variations in recall rates. We thus set $L$ to be 3 across all experiments.  
\begin{table*}[htbp]
	\caption{Effects of number of cross-attention fusion layers on recall rates.}
	\label{Tab:crossattentionlayer}
	\centering
	\resizebox{0.65\linewidth}{!}{
	\setlength{\tabcolsep}{1.5mm}{
	\begin{tabular}{c||ccc||ccc}
	\toprule
	\multirow{2}*{\#CA Layers $L$}  & \multicolumn{3}{c||}{MSLS-val}  & \multicolumn{3}{c}{SPED}\\
         \cline{2-7} 
        ~ & R@1 & R@5  & R@10 & R@1 & R@5  & R@10  \\
	\midrule 
     	 1              & 90.5 & 95.7 & 96.0          & 89.6 & 94.6 & 95.7 \\
	 3              & 92.0 & 96.4 & 96.9          & 89.8 & 94.7& 95.4 \\
	 5              &90.5 & 95.4 & 96.1           & 89.1 & 95.4 & 96.5 \\
	 7              &90.4 & 95.6 & 96.4           & 89.8 & 95.7 & 96.4 \\
       \bottomrule		 
      \end{tabular}}}
\end{table*}

\section{Performance Evaluation on Benchmark Nordland}
We also evaluate the performance of our proposed LVL3M on the challenging benchmark Nordland~\cite{Ali-Bey2023}, which was collected by recording a video from a train riding through the Norwegian countryside and traversing the same path across four seasons. The image are captured in suburban and natural environments, which typically lack salient objects for recognition. We follow MixVPR~\cite{Ali-Bey2023} to  use the summer and winter traverse as queries and reference dataset, respectively.  The recall rates are recorded in Table~\ref{Tab:nordland}. It can be seen that our LVL3M-VPR outperforms all other methods by a considerable margin except for CircaVPR.  Compared to the best-performing CircaVPR, our method brings an absolute decrease of 4.3\% at R@1. It is worth noting that although both CircaVPR and our LVL3M-VPR employ DINOv2 as the backbone for image feature extraction, CircaVPR develops an efficient adapter to improve the performance of pre-trained DINOv2 on VPR.  Since our main focus is on multi-modal fusion, we simply fine-tune the last four layers for domain adaption. This may explain why our performance is inferior to that of CircaVPR on Nordland. To further verify this speculation, we remove the proposed multi-feature combiner composed of text recalibration and cross-attention fusion components from our model and evaluate the performance of the reduced model, denoted as LVL3M-VP+ w/o CMB) on Nordland. It is observed that we only gain an R@1 of 53.0\%.  This demonstrate that our method could capture the useful information from noisy text descriptions, and benefits the learning of powerful feature representations on Nordland. 

\begin{table*}[htbp]
	\caption{Performance evaluation on benchmark Nordland. The best is highlighted in \textbf{bold} and the second is underlined. }
	\label{Tab:nordland}
	\centering
	\resizebox{0.65\linewidth}{!}{
	\setlength{\tabcolsep}{2.5mm}{
	\begin{tabular}{c||c||ccc}
	\toprule
	\multirow{2}*{Setting} & \multirow{2}*{Dim} & \multicolumn{3}{c}{Nordland} \\
         \cline{3-5} 
        ~ &~ & R@1 & R@5  & R@10 \\
	\midrule 
     	 GeM~\cite{Arandjelovic2016}                               & 2048                        & 20.8 & 33.3 & 40.0       \\
	 NetVLAD~\cite{Arandjelovic2016}                        & 32768                      & 32.6 & 47.1 & 53.3       \\
	 CosPlace~\cite{Berton2022}                                & 512                           & 34.4 & 49.9 & 56.5        \\
	 MixVPR~\cite{Ali-Bey2023}                                  & 4096                       & 58.4 & 74.6 & 80.0\\
	 EigenPlaces~\cite{Berton2023}                           & 2048                        & 54.4 & 68.8  & 74.1 \\
	 CircaVPR~\cite{Lu2024}                                      & 4096                        & \textbf{69.3}& \textbf{85.9}& \textbf{89.6} \\
	 LVL3M-VPR(ours)                                                &  $768*3$                  & \underline{65.0} & \underline{81.1} & \underline{85.9}   \\
	 \midrule 
	  LVL3M-VPR+w/o CMB                                 &  $768*3$                  & 53.0 (-12.0) & 69.6 (-11.5) & 74.5 (-11.4)   \\

       \bottomrule		 
      \end{tabular}}}
\end{table*}

\section{Visualization of Attention Maps from AT-REC}
\begin{figure*}[htbp]
        \centering
	\includegraphics[width = 1.0\textwidth]{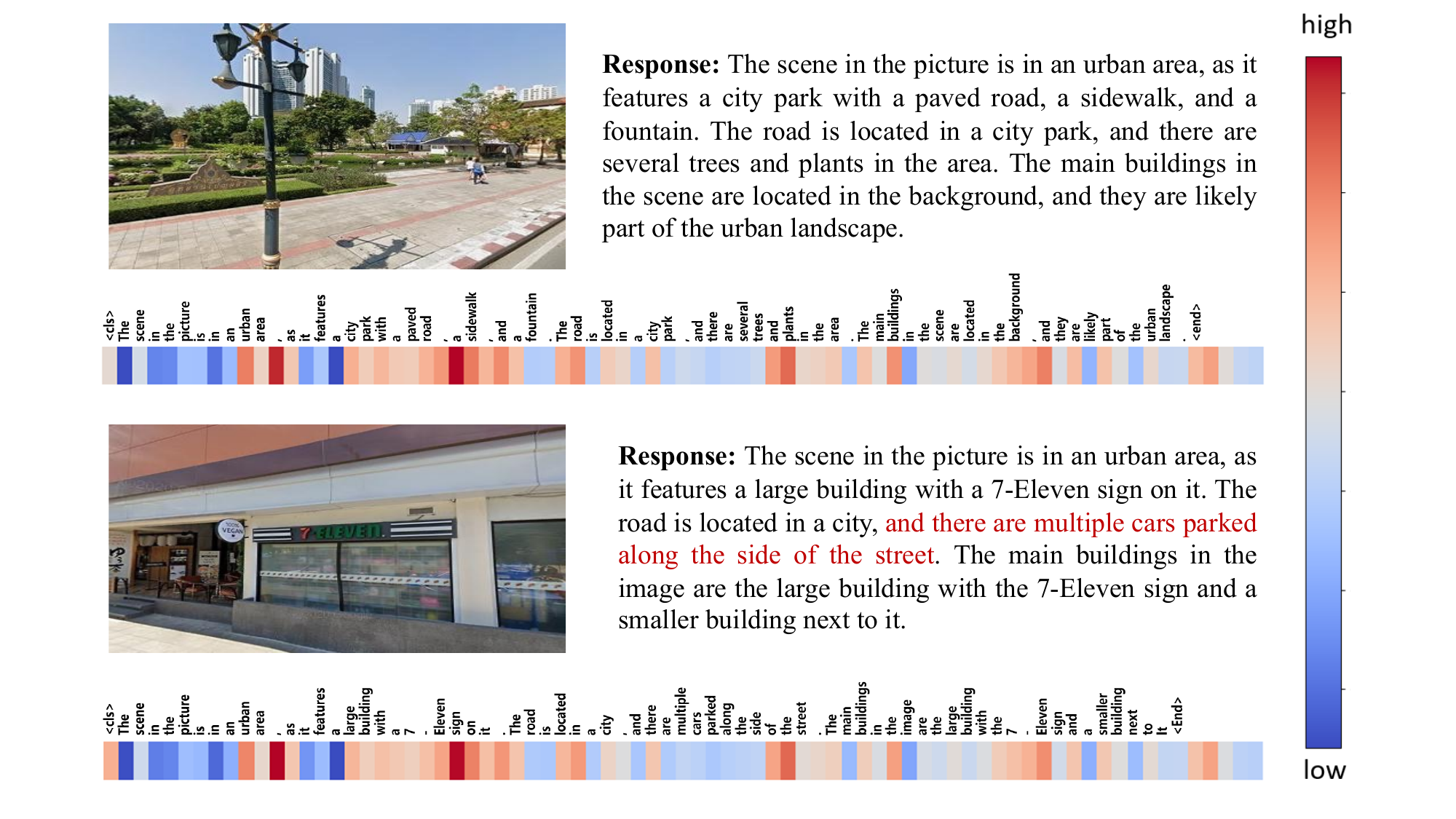}
	\caption{Visualization of the learned attention maps within AT-REC component. }
	\label{fig:Fig6-Attention-Recalibration}
\end{figure*}

We provide two examples to visualize the learned attention maps for recalibrating text sequences in Fig.~\ref{fig:Fig6-Attention-Recalibration}. For each example, we provide the raw image, the generated text description of the visual scene, as well as the generated attention map from AT-REC component. In the first example, the generated text description is accurate with the meaningful words such as ``urban'', ``city'', ``park'' , ``sidewalk'',  ``roads'', ``plants'' and ``buildings'' emphasized. For the second example, a majority of the description sentence is correct except the part marked in red. As can be seen, the proposed AT-REC assigns high weights to the words associated with the salient objects including ``7'', ``ELEVEN'', ``sign'' and ``buildings'' while those noisy and incorrect ones such as ``cars'' are depressed. The above observations demonstrate the effectiveness of our AT-REC in emphasizing informative words while depressing those noisy and incorrect ones. 

\section{Deep Insight into the Performance of LLaMA-Adapter~V2 on VPR}
\begin{figure*}{htbp}
	\centering
	\begin{subfigure}{0.99\linewidth}
		\includegraphics[width=1.0\linewidth]{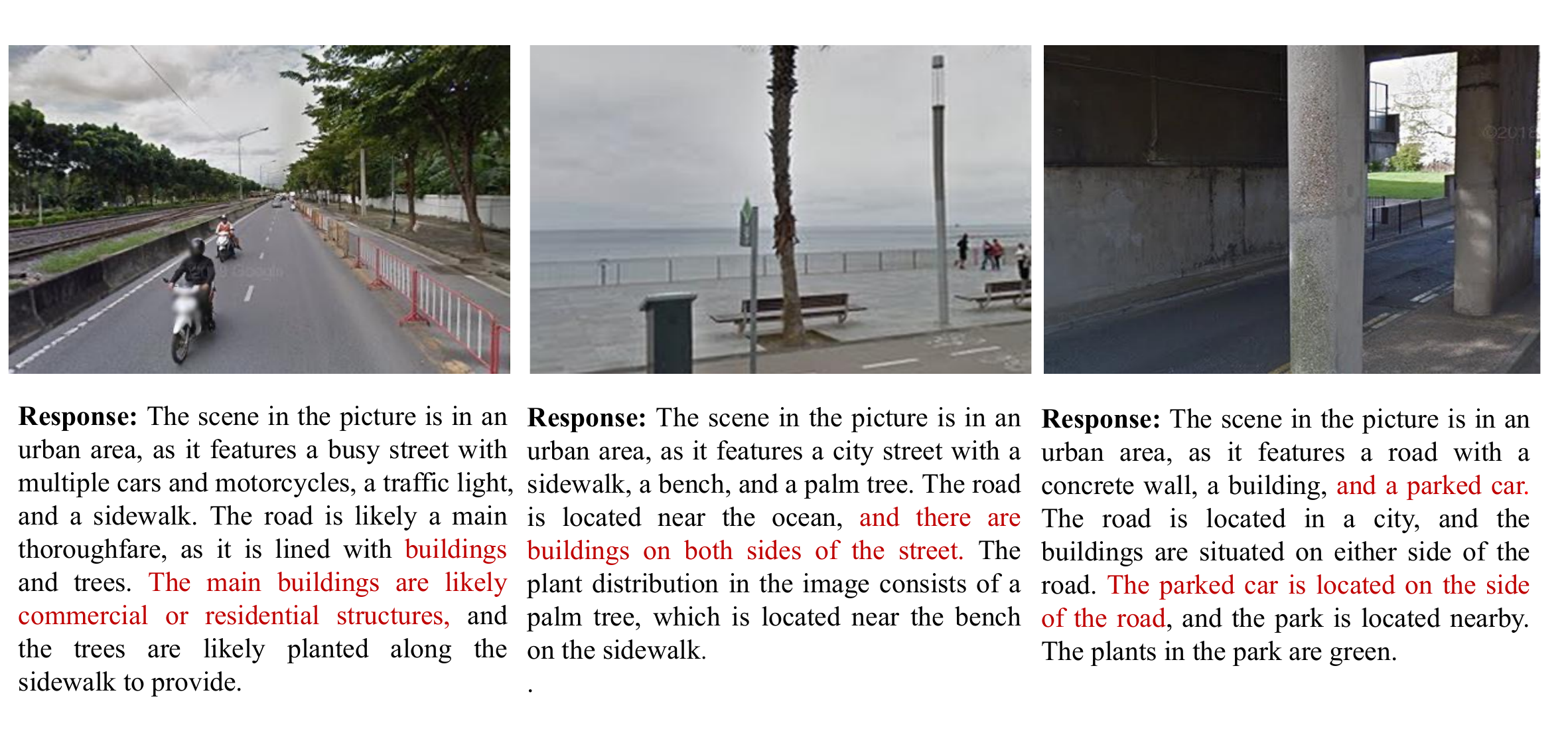}
		\caption{Generally accurate descriptions.}
		\label{fig:same-area}
	\end{subfigure}
	\begin{subfigure}{0.99\linewidth}
		\includegraphics[width=1.0\linewidth]{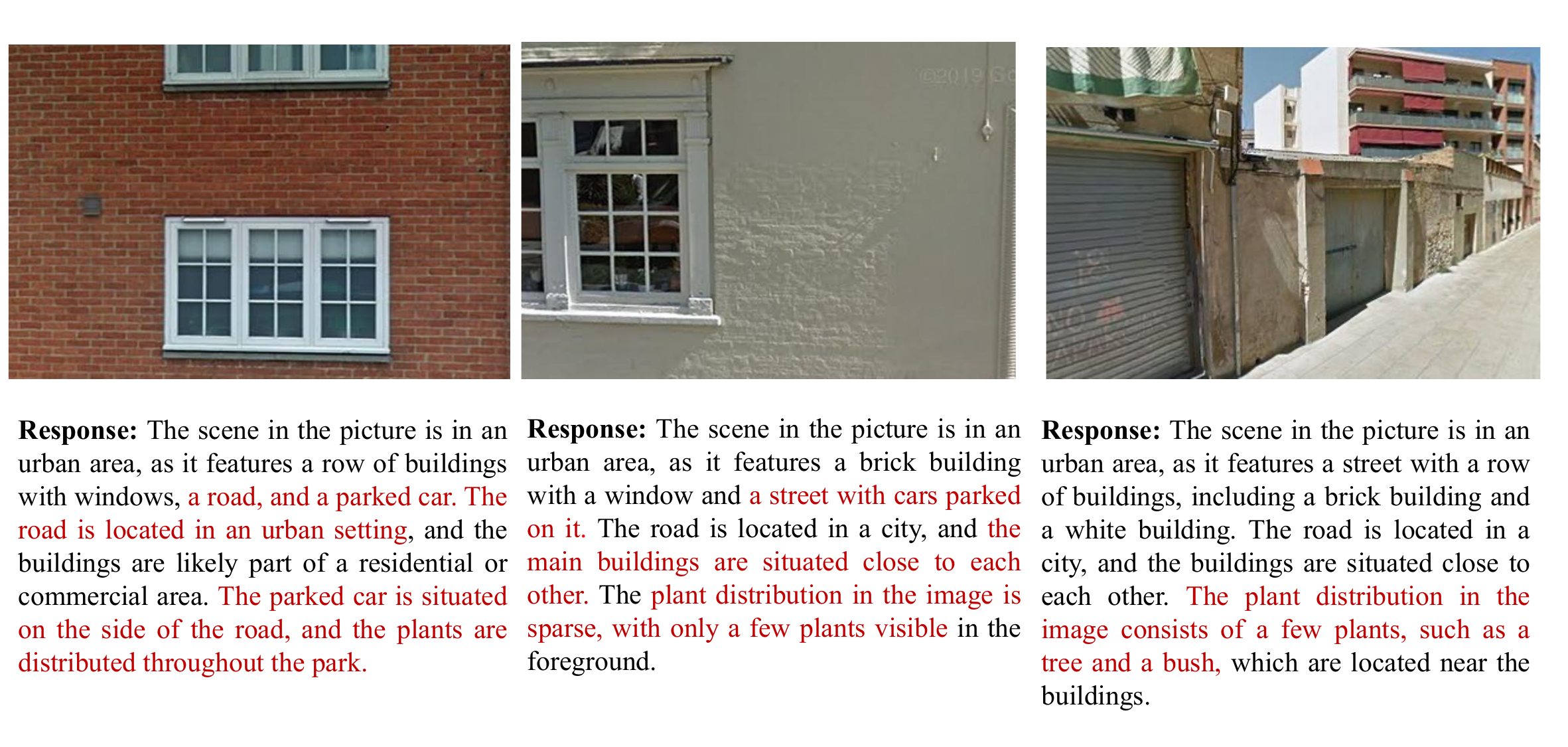}
		\caption{Noisy descriptions.}
		\label{fig:cross-area}
	\end{subfigure}
	\caption{ Performance of LLaMA-Adapter V2 on VPR. Inaccurate descriptions are marked in red.}
	\label{Fig:LLaMA-AdapterV2}
\end{figure*}

We provide several examples in Fig.~\ref{Fig:LLaMA-AdapterV2} to show the performance of the pre-trained LLaMA-Adapter~V2 on the downstream VPR task when presented with a prompt. For each example, we show the image and the generated text descriptions, with the incorrect descriptions marked in red. As can be seen, for the top three examples, the text descriptions generated by LLaMA-Adapter~V2 are generally accurate, with salient objects in the scene included in the description. However, in the three sample at bottom, the text descriptions generated by LLaMA-Adapter~V2 are noisy and incorrect, with some unseen objects included in the sentences.  These inaccurate descriptions easily confuse the proposed VPR model. The above observations suggest that although our multi-modal VPR already brings obvious improvement, its performance could be further enhanced by improving the performance of LVLM on VPR. This could be achieved by incorporating a more powerful LVLM or designing a more efficient prompt to cope with the high-diversity of VPR datasets. 

\end{document}